\documentclass{article}
\usepackage[preprint]{spconf}
\usepackage{amsmath,graphicx}

\usepackage[final]{listofsymbols}
\usepackage{xcolor}
\usepackage{hyperref}
\usepackage{cleveref}
\usepackage{mathtools}
\usepackage{multirow}
\usepackage{caption}
\usepackage{subcaption}
\usepackage{adjustbox}
\usepackage{lipsum}
\usepackage{booktabs}

\DeclareMathOperator*{\argmax}{arg\,max}

\makeatletter
\g@addto@macro\normalsize{%
	\setlength\abovedisplayskip{5pt plus 2pt minus 2pt}
	\setlength\belowdisplayskip{5pt plus 2pt minus 2pt}
	\setlength\abovedisplayshortskip{4pt plus 2pt minus 2pt}
	\setlength\belowdisplayshortskip{4pt plus 2pt minus 2pt}
}
\makeatother

\title{End-to-End Training of a Neural HMM\\with Label and Transition Probabilities}
\name{Daniel Mann$^{1}$, Tina Raissi$^{2}$,Wilfried Michel$^{1}$, Ralf Schl\"uter$^{1,2}$, Hermann Ney$^{1,2}$}

\address{
	$^1$AppTek GmbH, 52062 Aachen, Germany\\
	$^2$Machine Learning and Human Language Technology,\\
	Computer Science Department, RWTH Aachen University, 52074 Aachen, Germany\\
  \texttt{\{dmann,wmichel\}@apptek.com,
  \{raissi,schlueter,ney\}@cs.rwth-aachen.de}
}

\copyrightnotice{978-1-6654-7189-3/22/\$31.00~\copyright2023 IEEE}
\begin{document}
%
\maketitle

\opensymdef
\newsym[Silence symbol]{silence}{\texttt{sil}}
\newsym[Separator symbol]{sep}{\texttt{\#}}
\newsym[Set of all labels]{labelspace}{\Sigma}
\newsym[Set of all non-silence labels]{speech}{\bar\Sigma}
\newsym[Set of all states]{statespace}{S}
\newsym[Sequence of acoustic feature vectors]{features}{x_1^T}
\newsym[Sequence of hidden states]{states}{s_1^T}
\newsym[Sequence of labels]{phons}{a_1^S}
\newsym[Sequence of aligned labels]{alignment}{y_1^T}
\newsym[Phoneme pronounced like the english word ``I'']{ay}{\texttt{AY}}
\newsym[Phoneme pronounced like ``a'' in the english word ``am'']{aae}{\texttt{AE}}
\newsym[Phoneme pronounced like ``th'' in the english word ``the'']{tth}{\texttt{DH}}
\newsym[Training error criterion]{criterion}{\mathcal{L}}
\closesymdef
 
\begin{abstract}

We investigate a novel modeling approach for end-to-end neural network training using hidden Markov models (HMM) where the transition probabilities between hidden states are modeled and learned explicitly. Most contemporary sequence-to-sequence models allow for from-scratch training by summing over all possible label segmentations in a given topology. In our approach there are explicit, learnable probabilities for transitions between segments as opposed to a blank label that implicitly encodes duration statistics.

We implement a GPU-based forward-backward algorithm that enables the simultaneous training of label and transition probabilities.

We investigate recognition results and additionally Viterbi alignments of our models. We find that while the transition model training does not improve recognition performance, it has a positive impact on the alignment quality. The generated alignments are shown to be viable targets in state-of-the-art Viterbi trainings.

\end{abstract}

\section{Introduction}

Automatic speech recognition and hidden Markov models (HMM) enjoy a historical friendship. The task of speech recognition is not simple, requiring the mapping of a sequence of acoustic features to a sequence of words or phonemes. HMMs provided an effective solution to this problem by being a simple enough model to be efficiently computable and yet powerful enough to capture the temporal structure of speech. The essential modeling idea is to introduce a sequence of hidden states that has the same length as the sequence of acoustic features. It is supposed to capture the temporal structure of the speech signal -- at what time does the speaker pronounce which phoneme?


In their early applications in speech recognition, HMMs were generative models, i.e., modeling the distribution of the acoustic features, using Gaussian Mixture Models (GMM). Later, they were extended to discriminative models, so called discriminative HMMs \cite{bourlard1988links}, to take advantage of the enormous discriminative prowess of neural networks (NN). Today we find HMMs being used in the so called hybrid approach \cite{bourlard1994connectionist} still providing competitive results \cite{9054345,zeineldeen2022improving}. Before NN training in this approach a Viterbi alignment must be crafted by a GMM-HMM system. The NN is then trained to predict the Viterbi alignment with a cross-entropy loss on a framewise basis. Hence, if we think of end-to-end training, it is not HMMs that come to mind first but rather sequence-to-sequence models such as the Connectionist Temporal Classification (CTC) \cite{graves2006connectionist} or the Recurrent Neural Network Transducer (RNN-T) \cite{sak2017recurrent}.

But with this an important capability of the HMM approach is neglected. Only recently it has garnered some attention again \cite{zeyer2017ctc,raissi2023hmm} by using \textit{full-sum} HMM training which is closely related to CTC and RNN-T. In any full-sum training algorithm, the training loss is the weighted sum of the scores of different segmentations of the label sequence. For instance, in CTC one segmentation would be a possible padding of the label sequence with blank symbols. The weighting then depends on the output probability of the blank symbol. In HMMs, on the other hand, the weighting is given by the transition probabilities between hidden states.

This poses a big question: How do we determine these transition probabilities? While we could tune them or estimate them from given statistics, we may also learn them. In fact, transition model learning is a built-in mechanism of HMMs as part of the Baum-Welch algorithm \cite{rabiner1989tutorial} and is commonly used in other applications for HMMs such as computational biology \cite{krogh1994hidden} and computational finance \cite{mamon2007hidden}. The training of the transition model is so integral to HMMs in general, but virtually unexplored in the context of speech recognition.


 
It is for this reason that in this paper we investigate the role of transition model training in speech recognition. To achieve this we will revisit the forward-backward algorithm as it was originally envisioned since its introduction in \cite{baum1970maximization} -- as an efficient way to train both the transition model and the label posterior model.

Experimentally, we will find that the transition model training can be run in conjunction with the training of the label posterior. Although the WERs slightly degrade in comparison to the standard full-sum training of the label posterior alone, the quality of the alignments is improved by the addition of an explicit transition model. To our best knowledge, there is no well established measure for the evaluation of an alignment. We rely on a two-fold evaluation approach via (1) the ASR accuracy of a model trained on the alignment, and (2) the time stamp error (TSE) with respect to a GMM alignment as proposed in \cite{zhang2021lattice,chen2023improving,raissi2023hmm}. We then go on to show that our suggested training procedure can replace state-of-the-art CART-GMM pipelines for Viterbi training. Note that this is in stark contrast to CTC, where the alignments are known to exhibit peaky behavior \cite{graves2006connectionist,zeyer2017ctc}.

\section{Related work}



The training of both label and transition probabilities have been explored in so called Input-Output HMMs, where both models depend on another hidden state \cite{bengio1994input}. This is a generalization of classical HMMs capable of modeling the temporal structure of the transition model. However, the models had less expressive power and were trained with the EM-algorithm \cite{dempster1977maximum} as opposed to our fully neural approach.

More recently, conditional random fields (CRF) have appeared in automatic speech recognition in the form of hidden conditional random fields \cite{sung2009hidden}. These models are very similar to HMMs, in that it is possible to model label posterior and transition probability conjointly via a log-linear combination of a set of feature functions. The differences are that our training is inherently neural, which CRFs allow only with more effort \cite{artieres2010neural}, and the global normalization of CRFs differs from the local normalization of HMMs in its weaker constraint and thus higher computational cost. 

Full-sum HMM training with a fixed transition model has existed for a long while. For recent examples, see \cite{zeyer2017ctc,raissi2023hmm}. The training of an explicit neural HMM transition model is a novelty in speech recognition as far as the authors are aware.

The quality of alignments from different full-sum trainings has been studied extensively \cite{raissi2023hmm} and it is also not new to bootstrap Viterbi with an alignment from a full-sum trained model \cite{raissi2023competitive,zhou2022efficient}. Yet, the use of Viterbi trainings to compare different alignments is new to the authors.

\section{Models}

\subsection{Training Criterion}

Our goal is to model the posterior probability of a sequence of labels $\phons$ given the sequence of acoustic features $\features$. For this we will introduce a sequence of hidden states $\states$. This sequence has the same length as the feature sequence. It is a non-decreasing sequence of integers from $1$ to $S$.

We assume a first order dependence on $\states$. Thus we can write the sequence posterior probability as
\begin{align}
  p(\phons|\features) &= \sum_{\states} p(\phons, \states | \features) \notag \\
  &= \sum_{\states} \prod_{t} p_t(a_{s_t} | \features)\, p_t(s_{t} | a_{s_{t-1}}, s_{t-1},  \features). \label{eq:factorization}
\vspace{-1mm} 
\end{align}
Here we call $p_t(a | \features)$ the \textit{label posterior model} (LPM) and $p_t(s' | s, a, \features)$ the \textit{transition model} (TM). Note, that by choosing a "flat", that is, constant transition probability and replacing state repetitions with a blank label, we obtain the equation for the CTC loss \cite{graves2006connectionist}.

Our training objective function is the sequence posterior log-likelihood:
\vspace{-2mm}
\begin{equation*}
  \criterion = \log p(\phons|\features),
  \label{eq:loss}
\end{equation*}
which we will maximize with respect to the parameters of the label posterior and transition model via stochastic gradient descent. Furthermore, we will scale the label posterior and transition model with separate exponents.
  
\subsection{Gradients}

The gradients of the objective function are not new \cite{rabiner1989tutorial}. Yet the gradient with respect to the label posterior model has been discussed more frequently -- in the context of the CTC loss \cite{graves2006connectionist} and that of neural HMMs \cite{zeyer2017ctc}.

Let $\phons, \features$ be a label and a feature sequence, respectively. We introduce the following shorthand notation:
\begin{equation*}
  \Psi_{t,s}^{s'} = p_t(s'|s, a_{s}, \features), \qquad \Phi_t^s = p_t(a_{s} | \features).
\end{equation*}
\vspace*{-2mm}
Then we can write the label posterior gradient as
\begin{equation}
  \frac{\partial}{\partial \Phi_{t}^s} \criterion = \frac{\sum_{\states : s_t=s} \prod_t \Psi_{t,s}^{s'} \Phi_t^s }{\sum_{\states} \prod_t \Psi_{t,s}^{s'} \Phi_t^s }.
\end{equation}
For the transition model gradient let us first state the following auxiliary identity
\vspace{-2mm}
\begin{equation} \label{eq:aux}
  \frac{\partial}{\partial \Psi_{t,s}^{s'}} \prod_{t'} \Psi_{t',s_{t'-1}}^{s_{t'}} = \frac{\delta(s_{t-1},s) \delta(s_{t},s')}{\Psi_{t,s}^{s'}}  \prod_{t'} \Psi_{t',s_{t'-1}}^{s_{t'}}
\end{equation} 
for a given hidden state sequence $\states$.

Thus applying \eqref{eq:aux} to the gradient computation yields
\begin{align}
  \frac{\partial \log p(\phons|\features)}{\partial \log \Psi_{t,s}^{s'}} &= \frac{\Psi_{t,s}^{s'}}{p(\phons|\features)} \frac{\partial p(\phons|\features)}{\partial \Psi_{t,s}^{s'}} \notag \\
  &\mkern-25mu = \frac{\Psi_{t,s}^{s'}}{p(\phons|\features)} \sum_{\states} \big( \prod_{t'} \Phi_{t'}^{s_{t'}} \big) \frac{\partial}{\partial \Psi_{t,s}^{s'}} \prod_{t'} \Psi_{t',s_{t'-1}}^{s_{t'}} \notag \\
  &\mkern-25mu = \frac{1}{p(\phons|\features)} \sum_{\substack{\states\\s_{t-1}=s\\s_{t}=s'}} \prod_{t'} \Phi_{t'}^{s_{t'}} \Psi_{t',s_{t'-1}}^{s_{t'}}
\end{align}
This is the joint probability $p(s_{t-1}=s, s_{t}=s' | \phons, \features)$ of the pair of states $(s, s')$ occurring at times $t-1$ and $t$, respectively, in the sequence $\states$. 

The special case of a time-invariant and thus feature-independent transition model can be obtained by summing over all times frames $t$ in the above equation, that is,
\begin{equation}
  \frac{\partial \criterion}{\partial \log \Psi_{s}^{s'}} = \sum_t p(s_{t-1}=s, s_{t}=s' | \phons, \features).
\end{equation}
This can be interpreted as the expected number of times, the transition from $s$ to $s'$ occurs in the sequence $\states$.

\subsection{Transition Models}

For the label posterior model we use a combination of a deep neural network and a softmax layer to normalize the output. Primarily, we will use BiLSTM networks \cite{graves2005framewise}. We will call the network before the softmax layer the encoder.

The novelty in our modeling approach is the inclusion of an explicit, time-dependent transition model. Since the segmentation sequence $\states$ is a linearly non-decreasing sequence, only two probabilities are relevant for us: the \textit{forward probability} $p_F$ of moving to the next segment and the \textit{loop probability} $p_L$ of staying in the current segment. Of course, we impose the condition $p_F + p_L = 1$.

In general these two probabilities are functions of the input features $\features$ and the current time frame $t$, distinctive for each label $a$:
\vspace{-4mm}
\begin{equation}
  p_t(s'|s, a, \features) = \begin{cases}
    p_{F,a}(\features, t) & \text{if } s' = s + 1, \\
    p_{L,a}(\features, t) & \text{if } s' = s, \\
    0 & \text{otherwise}.
  \end{cases}
\end{equation}
For a time-invariant transition model the loop and forward probabilities are constants $p_{F,a}, p_{L,a}$ for each label $a$.

We will vary the degree of complexity of our transition models. In the simplest case we have a stationary transition model, where all speech states have the same transition probabilities $p_{F,a} = p_{F,\text{speech}}$ and only silence has a separate one $p_{F,\text{silence}}$. Adding complexity we present the following selection of models with their respective number of parameters:


\begin{itemize}
  \item \textbf{speech+silence} (2): as described above.
  \item \textbf{substate+silence} (4): Like speech+silence but distinuish between the three substate of speech states in tripartite HMM structure.
  \item \textbf{full} (250): Assign independent forward probabilities to each label and its substates.
  \item \textbf{full+input} (6.6M): Forward probabilities are output of a BiLSTM stacked on top of the encoder network. To the outputs a linear projection and a logistic sigmoid activation are applied. 
\end{itemize}


\subsection{Recognition}

Given a trained model we obtain the final word sequence via maximum a-posteriori rule, given the features $\features$, as follows
\begin{align} \label{eq:decision_rule}
  \resizebox{0.9\hsize}{!}{
    $\features \mapsto
    \underset{\phons}{\argmax}\hspace{-.4em} \left[ p^\beta \hspace{-.2em} (\phons) \max\limits_{\states} \prod\limits_{t} \frac{p(a_{s_t}| \features)}{p_{\text{prior}}^\gamma(a_{s_t})} p^\delta \hspace*{-.2em}  (s_{t}|s_{t-1}, a_{s_{t-1}}, \features) \right].$
  }
\end{align}

Here, $p(\phons)$ is the language model, and $p_{\text{prior}}(a)$ is the prior probability for each label $a$. The hyperparameters $\beta, \gamma, \delta$ are used to control the relative importance of the language model, the label prior and the transition model, respectively.

\section{Implementation}

The gradients are computed by a CUDA-implementation of the forward-backward algorithm. Our implementation is an extension of the label posterior gradient computation implemented by \cite{zeyer2017ctc} as part of the RETURNN framework \cite{doetsch2017returnn}.\footnote{Our code can be found at \url{https://github.com/DanEnergetics/returnn/tree/mann-fast-bw-tdps}}

The forward-backward algorithm is well know \cite{rabiner1989tutorial}. Essentially the goal is to compute the forward and backward probabilities
\begin{align*}
  \alpha_t(s) &= p(a_1^t, s_t = s | \features), \\
  \beta_t(s) &= p(a_{t+1}^T | s_t = s, a_s, \features)
\end{align*} 
which are computed via dynamic programming as shown in the following. We will only give a brief overview of the algorithm. The central quantities are the \textit{forward} and \textit{backward} probabilities:
\begin{align*}
  \alpha_t(s) &= p(a_1^t, s_t = s | \features), \\
  &= \sum_{\states: s_t = s} \prod_{t'=1}^{t} p_{t'}(a_{s_{t'}} | \features) p_{t'}(s_{t'}|s_{t'-1}, a_{s_{t'-1}}, \features) \\
  \beta_t(s) &= p(a_{t+1}^T | s_t = s, a_s, \features) \\
  &= \sum_{\states: s_{t} = s} \prod_{t'=t+1}^{T} p_{t'}(a_{s_{t'}} | \features) p_{t'}(s_{t'}|s_{t'-1}, a_{s_{t'-1}}, \features).
\end{align*} 
With the conditions $\alpha_1(s) = \delta(s, 1)$ and ${\beta_T(s)=\delta(s, S)}$, we can compute these probabilities recursively:
\begin{align*}
  \alpha_{t+1}(s) &= \sum_{s'} \alpha_t(s') p_{t+1}(a_{s'} | \features) p_{t+1}(s|s', a_{s'}, \features) \\
  \beta_{t}(s) &= \sum_{s'} \beta_{t+1}(s') p_{t+1}(a_{s'}| \features) p_{t+1}(s'|s, a_{s}, \features).
\end{align*}
To obtain the gradients let us define the following quantity:
\begin{equation*}
  \xi_t(s,s') \coloneqq \alpha_t(s) p_{t+1}(a_{s'}| \features) p_{t+1}(s'|s, a_{s}, \features) \beta_{t+1}(s').
\end{equation*}
From the definitions it is straightforward to identify
\begin{equation*}
  \xi_{t}(s,s') = p(s_{t}=s, s_{t+1}=s' | \phons, \features).
\end{equation*}
This is the gradient with respect to the time-dependent transition model. The gradient with respect to the label posterior model is obtained by summation over the successor state $s'$.

\section{Experiments}

\subsection{Setup}

Unless stated otherwise all label posterior models are 6-layer bidirectional LSTMs with 512 units each (34M parameters). The output is made into a probability distribution over the labels by a softmax layer. All weights and variables, including the transition model parameters, are optimized by the Adam optimizer with Nesterov accelerated gradient \cite{dozat2016incorporating}. The learning rate follows a cyclic schedule \cite{smith2017cyclical}. In particular, only one cycle is used where the minimal learning rate is $1.2 \times 10^{-5}$ and the maximal one is $3.0 \times 10^{-4}$. The cycle length is around $80 \%$ of training epochs and the learning rates are updated multiple times per epoch. We employ an L2 regularization for the weights with a scale of $10^{-4}$ and a dropout rate of $0.1$ \cite{srivastava2014dropout}.


For training and recognition we use the toolkits RETURNN \cite{doetsch2017returnn} and RASR~\cite{wiesler2014rasr}, respectively. For the training of the transition model we extended a pre-existent CUDA implementation of the forward-backward algorithm that computes the label posterior gradients.

We use two corpora: The LibriSpeech corpus (LBS) \cite{panayotov2015librispeech} and the Switchboard corpus (SWB) \cite{godfrey1992switchboard}. We train on (LBS: train-other, SWB: 300h subset) for (LBS: 25, SWB: 50) epochs and test on (LBS: dev-other, SWB: Hub5'00). As lexica we use the standard ones provided by the respective corpora and extend them with the OOV words via a grapheme-to-phoneme model trained on the corresponding lexicon. The target sentences are transduced to sequences of end-of-word monophones. We use a tripartite HMM topology with a monostate silence symbol.

For LBS and SWB we use 50 and 40 dimensional Gammatone features \cite{schluter2007gammatone}, respectively. All inputs to the network are augmented using SpecAugment \cite{park19e_interspeech}.


Lastly, if not stated otherwise, we use $0.3$ for transition and label posterior scale in training.

\subsection{Model initializations}

In order to test the stability and performance of the proposed training under different conditions we introduce different initialization strategies combining various initializations for the label posterior and the transition model.
\vspace*{0.5em}
\newline\noindent\textit{Transiton model:}
\begin{itemize}
  \item \textbf{Guessed}: We set $p_{F,\text{speech}} = 1/3$ for speech and $p_{F,\text{silence}}=1/40$ for silence. This is because these probability give exactly the average duration for pho\-nemes and silence in a tripartite model where we assume a geometrical distribution of the duration \cite{walck2007hand}.
  \item \textbf{Flat}: Uniform initialization $p_F = p_L = 1/2$.
  \item \textbf{Random}: Initialized according to the log-normal distribution \cite{atchison1980logistic}.
\end{itemize}
For the time-dependent case where the probabilities are functions of the input features, these initializations were used for the bias term in the topmost linear layer. The weight matrix is set to zero. \\
\newline\noindent\textit{Label posterior model:}
\vspace*{-0.7em}
\begin{itemize}
  \item \textbf{Random}: All appearing biases are initialized to zero while all forward weight matrices are initialized according to the Glorot uniform distribution \cite{glorot2010understanding}.
  \item \textbf{Pretrained}: Pretrain the model for 2 epochs with fixed, guessed transition model.
  \item \textbf{Pretrained+prior}: Pretrain the model as above but with a prior correction as in the classification rule \Cref{eq:decision_rule}.
\end{itemize}

\subsection{Training of the transition model}

In this section we investigate the effect of training the transition model on the performance of full-sum training. In \Cref{tab:corpus_tm} we compare label posterior training with and without transition model training on all corpora. The speech+silence and input dependent models gave the best WER results on SWB and LBS, respectively. The label posterior only training outperforms the joint training in terms of WER although the training criterion was more optimized or at the same level.

In \Cref{tab:best_init} we present the best WERs we obtained for each of the four proposed transition models across all different initialization strategies and the corresponding best initialization. We observe that a lower number of parameters in the transition model is rewarded with better WER results.

\begin{table}[t]
  \centering
  \caption{Comparison of transition model (TM) training and label posterior only training on different corpora. We took the best parametrizations for each corpus: speech+sil. on SWB and input dependent on LBS. Next to the WERs of each model are the training scores, i.e. the value of the training criterion $\criterion$, that the model reached and the time stamp error (TSE) of the alignment generated with the same model.}
  \label{tab:corpus_tm}
  \begin{tabular}{ccccc}
    \toprule
    corpus & TM training & WER [\%] & Score & TSE \\ 
    \midrule
    \multirow{2}{*}{SWB} & no & 13.8 & 0.06 & {4.7} \\
    {} & yes & 14.0 & 0.6 & {5.0} \\
    \midrule
    \multirow{2}{*}{LBS} & no & 8.8 & 0.08 & {7.1} \\
    {} & yes & 12.9 & 0.08 & {4.7} \\
    \bottomrule
  \end{tabular}
\end{table}

\begin{table}[t]
  \centering
  \caption{Best combination of initialization strategies for transition model (TM) and label posterior model (LPM) for each transition model parametrization on SWB. Next to it is the corresponding minimal word error rate (WER).}
  \label{tab:best_init}
  \begin{tabular}{cccc}
    \toprule
    \multirow{2}{*}{Param.} & \multicolumn{2}{c}{Best Init.} & \multirow{2}{*}{WER [\%]} \\
    {} & TM & LPM & {} \\
    \midrule
    speech+sil. & flat & random & 14.0 \\
    substate+sil. & guess & pretrain+prior & 14.1 \\
    full & flat & random & 14.4 \\
    full+input & guess & pretrain+prior & 14.9 \\
    \bottomrule
  \end{tabular}
\end{table}

The learned probability parameters were close to the guessed values with high consistency. For this we read out the learned transition probabilities from the trained models without input dependence. The speech forward probability averaged around $0.41$ and the silence forward probability around $0.22$. Compare this to the guessed values of $1/3 \approx 0.33$ and $1/40 = 0.025$, respectively.

\subsection{Alignments}

We saw no word error rate improvement although transition model training improved the training scores (see \Cref{tab:corpus_tm}). But what if we have looked at the wrong use case? The forced alignment task is much more closely related to our training criterion. In this chapter we will look at the quality of the alignments generated by the different transition models. 

To measure the quality of the alignments we looked at two metrics. The first one is the time stamp error (TSE) which was introduced in \cite{zhang2021lattice}.
While in the original paper it is the mean absolute displacement between start/end time-stamps of words hypothesized in decoding, we consider this displacement on the phoneme level between two alignments. Here we take only those sentences with equal pronunciations in both alignments. In our evaluation the reference alignment was generated by a triphone GMM-HMM as GMMs are still the standard technology for the production of initial alignments.

The second metric is the word error rate of a Viterbi training that uses the alignment generated by the given model.

\Cref{tab:alignments_tdp} shows that the alignments generated by the full-sum trained models are of high quality. The TSE is very low and the Viterbi training using the generated alignments yields a WER that is very close to the WER of the full-sum trained model itself. Furthermore, the learned full transition model managed to outperform the fixed transition model.

\begin{figure}[t]
  \begin{subfigure}[b]{0.463\textwidth}
    \centering
    \hspace*{-0.5cm}
    \includegraphics[width=\textwidth]{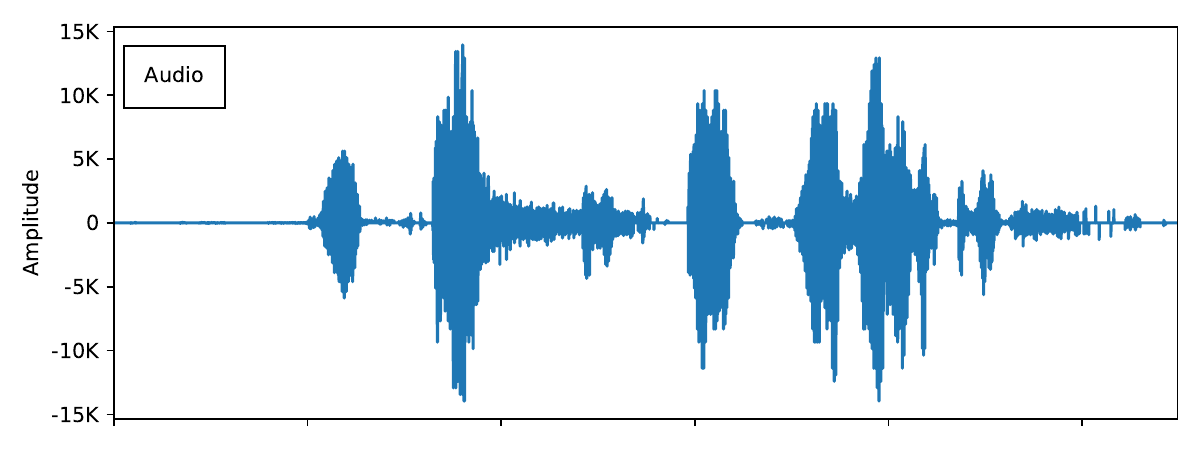}
  \end{subfigure}

  \vspace{-.2cm}

  \begin{subfigure}[b]{0.45\textwidth}
    \centering
    \includegraphics[width=\textwidth]{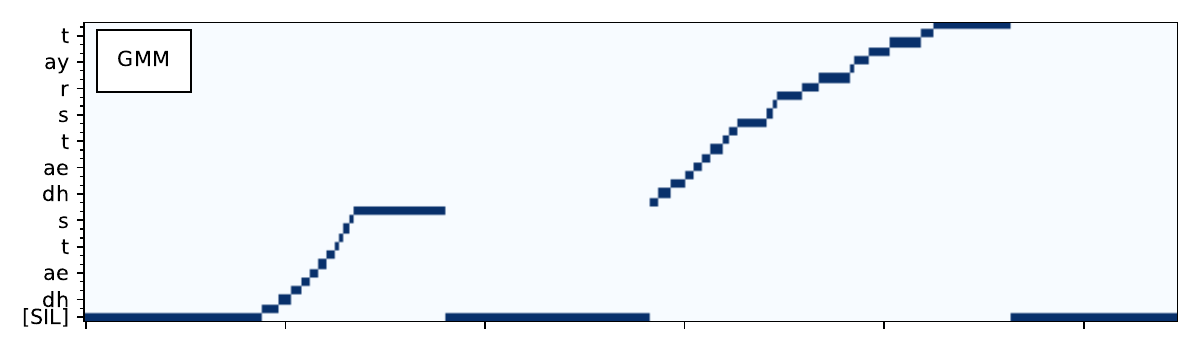}
  \end{subfigure}

  \vspace{-.2cm}

  \begin{subfigure}[b]{0.45\textwidth}
    \centering
    \includegraphics[width=\textwidth]{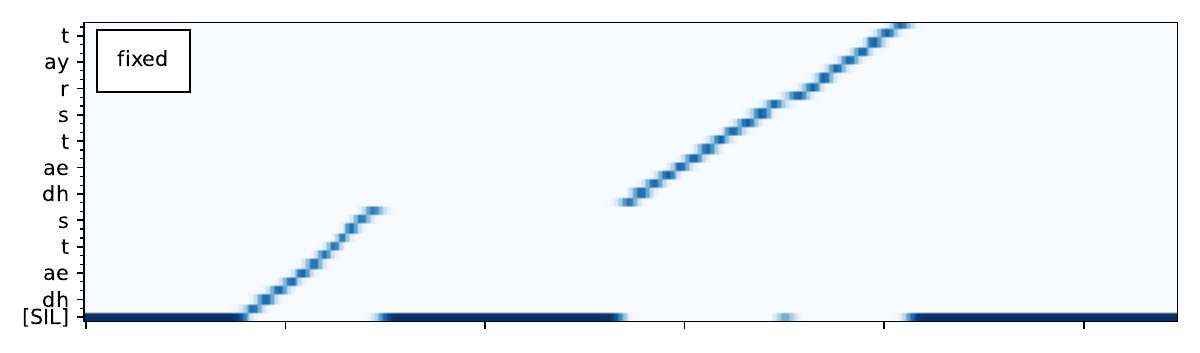}
  \end{subfigure}
  
  \vspace{-.2cm}



  \begin{subfigure}[b]{0.45\textwidth}
    \centering
    \includegraphics[width=\textwidth]{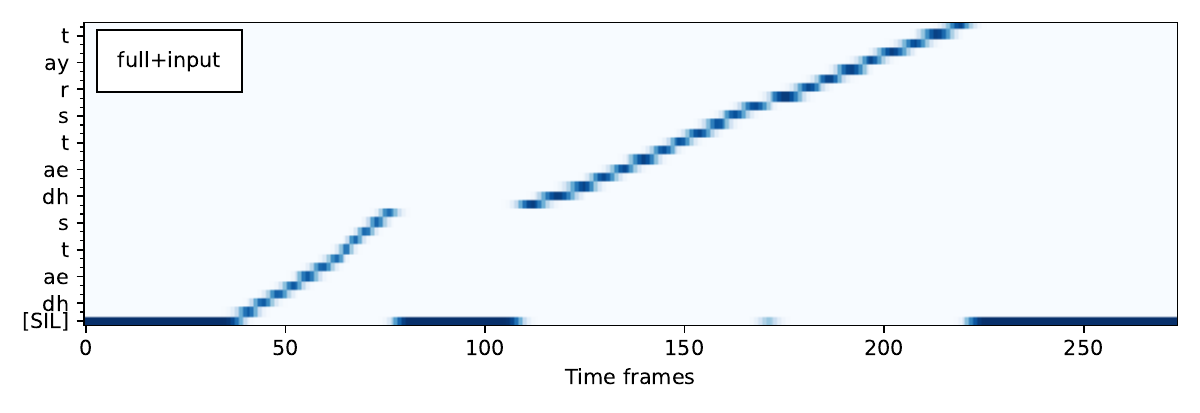}
  \end{subfigure}

  \vspace{-.2cm}


  \caption{Soft alignments for full-sum models, forced alignment for GMM and the audio signal of the utterance "That's that's right". Note the louder segment from frame 80 to 140 is the laughter of the speaker. Since the corpus transcription does not contain a laughter symbol here, the models chooses silence instead. Each image is labeled with its encoder or transition model type.}
  \label{fig:alignments}
\vspace{-5mm}
\end{figure}

\Cref{fig:alignments} clarifies the differences between the alignments generated by the different models. The BiLSTM alignments show very linear slopes. It shows virtually none of the variation in duration that is present in the GMM alignment.

\begin{table}[t]
  \centering
  \caption{Comparison of alignments for trained/fixed transition models and different transition model architectures on SWB. Next to the WER performance of the model itself, we show the time stamp error (TSE) of the alignment generated by the model. The last column shows the WER of a Viterbi-BiLSTM training that uses the alignment generated by the full-sum model. The Viterbi training is configured identically to the full-sum training.}
  \label{tab:alignments_tdp}
  \begin{tabular}{ccccc}
      \toprule
      \multicolumn{2}{c}{Transition model} & \multirow{2}{*}{TSE} & {Viterbi} \\
      param. & trained & & WER [\%] \\
      \midrule
      speech+silence & no & {4.7} & {14.6} \\
      \midrule
      speech+silence    & \multirow{4}{*}{yes} & {5.0} & {14.2}\\
      substate+silence  &                      & {4.9} & {14.2}\\
      full              &                      & {4.7} & {14.4}\\
      full+input        &                      & {5.0} & {14.5}\\
      \bottomrule
  \end{tabular}
\end{table}

\begin{table}[t]
  \centering
  \caption{Viterbi training performance of our full transition model alignment and a state-of-the-art alignment on SWB. The Viterbi training was conducted with factored hybrid (FH) and CART state-tying. The reference alignment is produced with a GMM-NN system.}
  \label{tab:best}
  \vspace{0.1cm}
  \begin{tabular}{ccc}
    \toprule
    Alignment & Tying & Viterbi WER [\%] \\
    \midrule
    \multirow{2}{*}{GMM-NN} & CART & {12.9} \\
     & FH & {12.7} \\
    \midrule
    This work & FH & {12.7} \\
    \bottomrule
  \end{tabular}

\end{table}

Lastly, we tested the alignment of the full transition model against a high quality alignment known to yield state-of-the-art performance in a conformer hybrid ASR setting \cite{zeineldeen2022improving}. This alignment was generated by a speaker-adaptive tandem GMM-NN system as in \cite{tuske2015speaker}. For the Viterbi training we used a triphone CART system and a triphone factored hybrid \cite{raissi2020fh} approach where the context is modeled explicitly via phoneme embeddings. It thus obviates the need for an additional clustering algorithm.  The training settings are identical to the full-sum training apart from a peak learning rate of $8 \times 10^{-4}$. This fully neural pipeline produces competitve results (see \Cref{tab:best}).

\begin{table}[t]
    \centering
    \caption{Comparison of alignments for different label posterior models on SWB. Next to the WER performance of the model itself, we show the TSE of the alignment generated by the model. The last column shows the WER of a Viterbi training that uses the alignment generated by the full-sum model.}
    \label{tab:encoder_alignments}
    \vspace{0.1cm}
    \begin{tabular}{ccc}
        \toprule
        LPM     & TSE & Viterbi WER [\%] \\
        \midrule
        BiLSTM  & {5.0} & {16.5} \\
        {GMM}   & (0.0) & {16.6} \\
        {FFNN}  & 2.3 & {16.5} \\
        {TDNN}  & 2.6 & {16.0} \\
        \bottomrule
    \end{tabular}
\end{table}

\subsection{Impact of the encoder architecture}


We chose BiLSTMs as our encoder for a better conceptual comparison with a modern CTC pipeline and a good trade-off between direct WERs of the model and alignment quality. However, we want to point out that the alignments can be further improved by employing different encoders. For this we showed alignments produced by a 6-layer feed forward neural network (FFNN) and a 8-layer time-delay neural network (TDNN) in \Cref{tab:encoder_alignments}. Both of these models did not undergo any particular tuning and we did not employ transition model training. This hints into a possible direction, in case even better alignments are desired at the cost of direct WERs.

\subsection{Impact of Scales}

\begin{table}[t]
  \centering
  \caption{Comparison of different scale settings for full-sum training with and without transition model (TM) training on SWB. Here we show the WER of the model itself and the TSE of the alignment generated by the model. We use the speech+silence transition model and scale it by $\{0.1, 0.3\}$ in the training criterion.}
  \label{tab:scales}
  \vspace{0.2cm}
  \begin{tabular}{cccc}
      \toprule
      {Train TM} & {TM scale} & {WER [\%]} & {TSE} \\
      \midrule
      \multirow{2}{*}{no} & 0.1 & {13.8} & {4.7} \\
      {} & 0.3 & {13.9} & {9.5} \\
      \midrule
      \multirow{2}{*}{yes} & 0.1 & {14.0} & {5.0} \\
      {} & 0.3 & {14.9} & {6.9} \\
      \bottomrule
  \end{tabular}
\end{table}

The success of full-sum training seems to heavily rely on the choice of the log-linear scales in the training criterion. Indeed, already small deviations from our baseline scales $(0.3, 0.3)$ for label posterior and transition model could lead to a significant drop in performance or even break the model. Thus, under the natural choice of scales $(1.0, 1.0)$ the training did not converge at all.
In \Cref{tab:scales} we see that a small change of transition model scale did not influence the WER while more heavily impacting the transition model training. The alignment quality deteriorated significantly with the lower scale for both cases.

\section{Discussion}


Our experiments show, that the conjoint training of label posterior and transition model has slightly worse recognition performance than the training of the label posterior with a fixed TM. Comparing different parametrizations for the transition model, we could observe a tendency towards a lower numbers of parameters.


Nonetheless, the proposed training improved the alignment quality compared to the training with fixed TM. Moreover, we were able to obtain state-of-the-art Viterbi training results by using the alignments generated by the full-sum model. This shows that the proposed training strategy is a viable alternative to the standard Viterbi training pipeline where usually multiple GMM iterations are used to generate the initial alignment. Our pipeline needs only one training run and requires no prior knowledge like phoneme duration statistics as they are learned during the training.


\section{Conclusions}


With our work, we hope to make another step towards rekindling interest in HMMs in automatic speech recognition.\ Similarly to other sequence-to-sequence approaches, HMMs with label posterior and different variants of a transition model can be trained in an end-to-end neural fashion and generate meaningful alignments in the process. Such models can be seen as a special case of CTC without the blank symbol in the label set, and with an explicit transition model. We show that while the joint training of label and transition model does not improve the recognition performance, it helps generating better alignments.

Moreover, current state-of-the-art hybrid models rely on GMM based systems within a heterogeneous pipeline consisting of different training criteria in order to obtain the alignment for the frame-wise cross-entropy training. We showed that full-sum neural HMM training can replace this step and provide alignments of similar if not better quality with the help of transition model training. In the future, we will further investigate the potential of this approach by using other encoder architectures, and by exploring alternative optimization strategies for the transition model.





\section{Acknowledgements}

This work was partially supported by NeuroSys, which as part of the initiative “Clusters4Future” is funded by the Federal Ministry of Education and Research BMBF (03ZU1106DA).

\bibliographystyle{IEEEbib-abbr}
\bibliography{refs}

\end{document}